\documentclass[conference]{IEEEtran}
\IEEEoverridecommandlockouts

\usepackage{cite}
\usepackage{amsmath,amssymb,amsfonts}
\usepackage{algorithmic}
\usepackage{graphicx}
\usepackage{textcomp}
\usepackage{xcolor}
\usepackage{booktabs}
\usepackage{url}
\def\BibTeX{{\rm B\kern-.05em{\sc i\kern-.025em b}\kern-.08em
		T\kern-.1667em\lower.7ex\hbox{E}\kern-.125emX}}

\makeatletter
\newcommand{\linebreakand}{%
\end{@IEEEauthorhalign}
\hfill\mbox{}\par
\mbox{}\hfill\begin{@IEEEauthorhalign}
}
\makeatother

\begin{document}
	
	\title{Multi-Objective Large Language Model Unlearning\\
		\thanks{$^*$Zibin Pan and Shuwen Zhang contributed equally to the paper.}
		\thanks{$^\dag$Junhua Zhao is the corresponding author.}
		\thanks{CUHKSZ: The Chinese University of Hong Kong, Shenzhen, China.}
		\thanks{Zibin Pan is also in The Cyberspace Academy of Guangzhou University.}
		\thanks{\textcopyright ~2025 IEEE. Personal use of this material is permitted. Permission from IEEE must be obtained for all other uses, in any current or future media, including reprinting/republishing this material for advertising or promotional purposes, creating new collective works, for resale or redistribution to servers or lists, or reuse of any copyrighted component of this work in other works.}
	}
	
	\author{\IEEEauthorblockN{1\textsuperscript{st} Zibin Pan$^*$}
		\IEEEauthorblockA{\textit{School of Science and Engineering} \\
			\textit{CUHKSZ}\\
			{Shenzhen, China} \\
			zibinpan@link.cuhk.edu.cn}
		\and
		\IEEEauthorblockN{2\textsuperscript{nd} Shuwen Zhang$^*$}
		\IEEEauthorblockA{\textit{School of Science and Engineering} \\
			\textit{CUHKSZ}\\
			{Shenzhen, China} \\
			shuwenzhang@link.cuhk.edu.cn}
		
		\and
		
		\IEEEauthorblockN{3\textsuperscript{rd} Yuesheng Zheng}
		\IEEEauthorblockA{\textit{School of Science and Engineering} \\
			\textit{CUHKSZ}\\
			{Shenzhen, China} \\
			yueshenzheng@link.cuhk.edu.cn}
		
		\linebreakand
		
		\IEEEauthorblockN{4\textsuperscript{th} Chi Li}
		\IEEEauthorblockA{\textit{School of Data Science} \\
			\textit{CUHKSZ}\\
			{Shenzhen, China} \\
			chili@link.cuhk.edu.cn}
		
		\and
		
		\IEEEauthorblockN{5\textsuperscript{th} Yuheng Cheng}
		\IEEEauthorblockA{\textit{School of Science and Engineering} \\
			\textit{CUHKSZ}\\
			{Shenzhen, China} \\
			yuhengcheng@link.cuhk.edu.cn}
		\and
		\IEEEauthorblockN{6\textsuperscript{th} Junhua Zhao$^\dag$}
		\IEEEauthorblockA{\textit{School of Science and Engineering} \\
			\textit{CUHKSZ}\\
			{Shenzhen, China} \\
			zhaojunhua@cuhk.edu.cn}
	}
	
	\maketitle
	
	\begin{abstract}
		Machine unlearning in the domain of large language models (LLMs) has attracted great attention recently, which aims to effectively eliminate undesirable behaviors from LLMs without full retraining from scratch. In this paper, we explore the Gradient Ascent (GA) approach in LLM unlearning, which is a proactive way to decrease the prediction probability of the model on the target data in order to remove their influence. We analyze two challenges that render the process impractical: gradient explosion and catastrophic forgetting. To address these issues, we propose Multi-Objective Large Language Model Unlearning (MOLLM) algorithm. We first formulate LLM unlearning as a multi-objective optimization problem, in which the cross-entropy loss is modified to the unlearning version to overcome the gradient explosion issue. A common descent update direction is then calculated, which enables the model to forget the target data while preserving the utility of the LLM. Our empirical results verify that MoLLM outperforms the SOTA GA-based LLM unlearning methods in terms of unlearning effect and model utility preservation. The source code is available at \url{https://github.com/zibinpan/MOLLM}.
	\end{abstract}
	
	
	\section{Introduction}
	\label{sec:intro}
	
	The wide application of LLM in various fields raises significant concerns related to its safety, including but not limited to harmful responses \cite{bai2022constitutionalaiharmlessnessai}, copyright infringement \cite{meeus2024copyrighttrapslargelanguage}, hallucinations \cite{ji2023towards}, and user privacy \cite{sebastian2023privacy, huang2023firewallm}. An intuitive solution to address these concerns would be to retrain the model on an updated dataset that excludes undesirable samples. However, such drastic measure is both time-consuming and financially prohibitive \cite{yao2023large,liu2024rethinking}. Machine unlearning, in contrast, offers an efficient and cost-effective alternative by enabling the selective removal of data, allowing the LLM to ``forget" specific information \cite{xing2024efufefficientfinegrainedunlearning, dou2024avoidingcopyrightinfringementmachine, chen2023unlearn}.
	
	Numerous LLM unlearning techniques have been proposed recently \cite{liu2024rethinking}. For instance, \cite{pawelczyk2023context} treats LLMs as a black box, and proposes an in-context unlearning method. However, the undesirable behaviors still persist within the model, and prompt manipulation can cause the model to output undesirable information \cite{liu2023formalizing}. Apart from that, Eldan and Russinovich \cite{eldan2023whosharrypotterapproximate} propose a relabeling fine-tuning method to unlearn the target information in LLM, where the label of the undesirable data is altered before fine-tuning the LLM on the updated dataset. RLHF (reinforcement learning from human feedback) and its variants \cite{ouyang2022training,lee2023rlaif} utilize human-written outputs and fine-tuning to calibrate the responses of LLMs. Unfortunately, these methods are resource-intensive and computationally costly \cite{liu2024rethinking}. In resource-constrained scenarios, removing harmful responses should be prioritized over generating desirable responses \cite{yao2023large}. To this end, Gradient Ascent (GA) method that increases the training loss $\mathcal{L}_{fgt}$ on the target forget data (i.e., the data that requires to be forgotten) to achieve unlearning \cite{yao2023large}, offers a resource-efficient and proactive alternative to complete retraining. Notably, GA has become a significant branch in the field of machine unlearning \cite{trippa2024nablataugradientbasedtaskagnostic,zagardo2024more}. However, two major challenges arise when adopting GA in LLM unlearning: gradient explosion and catastrophic forgetting.
	
	\textbf{Gradient explosion}. Since the Cross-Entropy (CE) loss function has no upper bound, adopting GA to unlearn the target information in LLM would increase the gradient without bound and even lead to gradient explosion. To tackle this issue, one naive way is the gradient clipping method \cite{halimi2023federatedunlearningefficientlyerase}, which limits gradient norms with an extra hyper-parameter. Nevertheless, experimental tuning is required to find the optimal hyper-parameter. In contrast, we introduce an effective solution that replaces the Cross-Entropy loss with its unlearning version, and uses gradient descent to achieve the unlearning goal. In this way, the issue of gradient explosion could be overcome without the need of tuning extra hyper-parameters.
	
	\begin{figure}[t!]
		\centering
		\includegraphics[width=0.95\columnwidth]{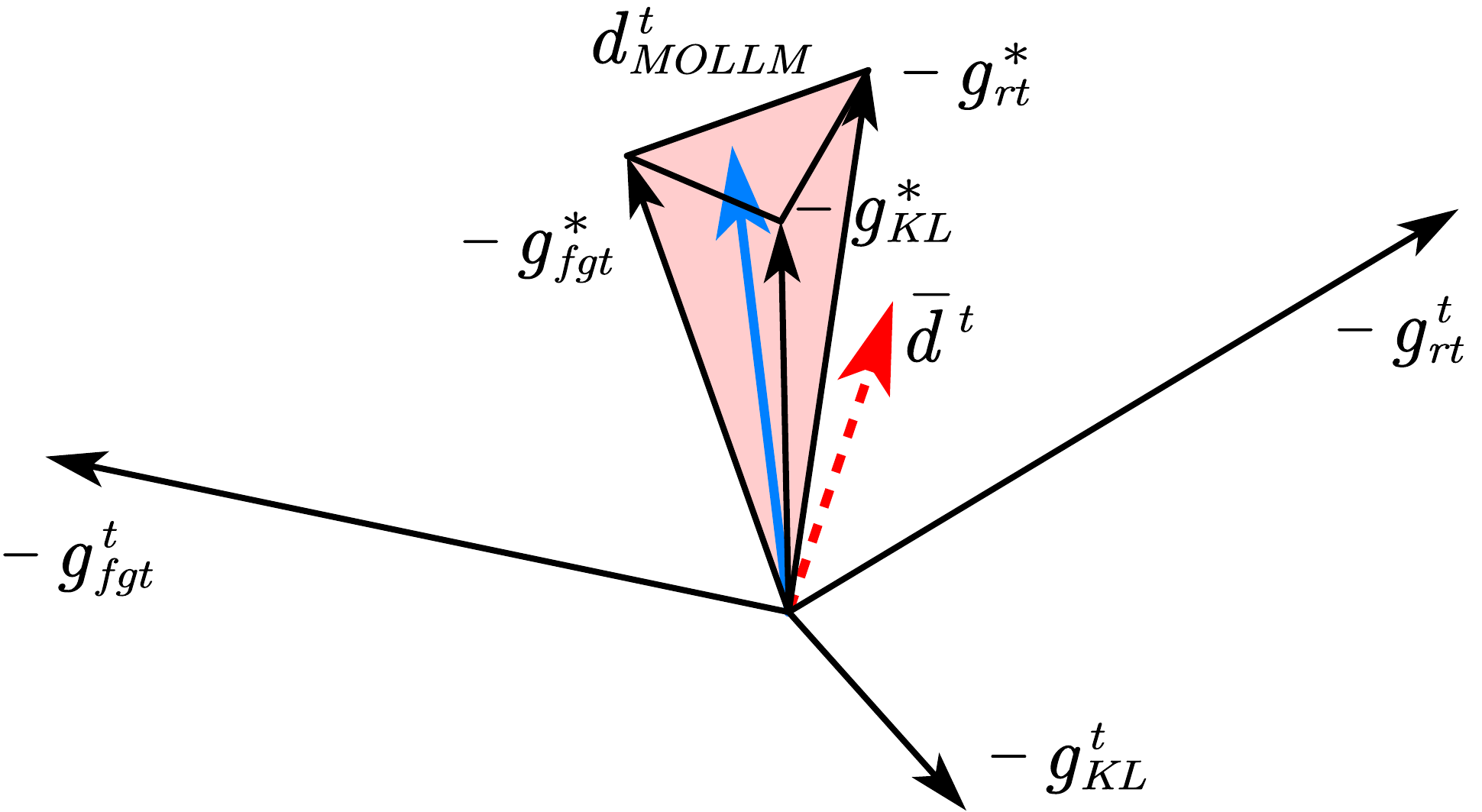}
		\caption{A demo of the gradient conflict during LLM unlearning, where $g_{fgt}$, $g_{KL}$, $g_{rt}$ denote the gradients of $\mathcal{L}_{fgt}$, $\mathcal{L}_{KL}$, and $\mathcal{L}_{rt}$, respectively. $\mathcal{L}_{fgt}^*$, $\mathcal{L}_{KL}^*$, and $\mathcal{L}_{rt}^*$ are the corresponding dual vectors. $\bar d^t$ represents the update direction of the weighted-sum method, which takes the weighted-sum of $-g_{fgt}$, $-g_{KL}$, and $-g_{rt}$. The hull in the pink color denotes the dual space of span($-g_{fgt}$, $-g_{KL}$, $-g_{rt}$), where all the vectors inside are the common descent directions. The blue direction $d^t_{MOLLM}$ is the common descent direction obtained by the proposed MOLLM.}
		\label{fig:fig1}
	\end{figure}
	
	\textbf{Catastrophic forgetting in downstream tasks.} This challenge is not unique to GA-based approaches but is common across fine-tuning-based LLM unlearning methods. In LLM unlearning, the goal is twofold: not only to unlearn the target forget data, i.e., reducing the model performance on them, but also to maintain the model utility (i.e., performance) on downstream tasks \cite{yao2023large, chen2023unlearn, kurmanji2023unboundedmachineunlearning}. To address this, Yao et al. \cite{yao2023large} introduce $\mathcal{L}_{KL}$, the Kullback-Leibler (KL) divergence between the outputs of the unlearned and the original models on the retain data, and incorporate it into the unlearning loss function by the weighted-sum method. Furthermore, \cite{chen2023unlearn} add the Cross-Entropy loss $\mathcal{L}_{rt}$ of the model on the retain data to maintain the model's performance on downstream tasks. 
	
	However, simply aggregating the unlearning loss, the KL loss, and the loss on the retain data cannot resolve the conflict between unlearning efficacy and the preservation of model utility. This is because the gradient of the unlearning process easily conflicts with the gradients of the KL loss and $\mathcal{L}_{rt}$. Hence, the naively-aggregated direction could conflict with the gradient of $\mathcal{L}_{fgt}$, $\mathcal{L}_{KL}$, or $\mathcal{L}_{rt}$ (as seen in Fig.~\ref{fig:fig1}). We verify this conflict in the experimental results in Table~\ref{tab:exps}. 
	
	
	To address this issue, we propose the \textbf{M}ulti-\textbf{O}bjective \textbf{LLM} unlearning (MOLLM) algorithm that computes a common descent direction for the model update, which can reduce the unlearning loss while preserving the model utility.
	
	We summarize our contributions as follows:
	\begin{itemize}
		\item We introduce an unlearning version of Cross-Entropy loss to overcome the gradient explosion in LLM unlearning.
		\item We formulate LLM unlearning as a multi-objective optimization problem and propose a Dual Space Multiple Gradient Descent Algorithm (DS-MGDA) to calculate a common descent update direction, allowing the model to unlearn the target data while preserving the utility.
		\item We have validated our method on the SafeRLHF Dataset \cite{ji2024pkusaferlhfsafetyalignmentpreference}. The results confirm the superiority of our method in balancing unlearning efficiency with utility retention.
	\end{itemize}
	
	\section{Preliminaries}
	\subsection{Unlearning Settings and Goal}
	We follow the definition of LLM unlearning in \cite{yao2023large,chen2023unlearn}. Suppose an LLM with parameters $\theta$ has been trained to convergence on the training data $D_{tr}$ for a specific downstream task. Following the deployment of the model, some undesirable samples $D_{fgt} \subset D_{tr}$ are identified and need to be unlearned, while the model performance on the retain set $D_{rt} \subset D_{tr}$ should remain intact, with $D_{rt} \cup D_{fgt} = D_{tr}$. Unlearning is thus defined as a process that produces a new model that behaves as if it has never encountered $D_{fgt}$, while maintaining its utility on the retain set $D_{rt}$.

	\subsection{Catastrophic Forgetting in LLM Unlearning}
	GA is a proactive way to unlearn $D_{fgt}$, which takes the inverse update of learning to maximize the model loss on the forget set $D_{fgt}$. It could be expressed as:
	\begin{equation}
		\label{bad_loss}
		\mathcal{L}_{fgt} := - \sum_{(x^{fgt}, y^{fgt}) \in D_{fgt}} L(x^{fgt}, y^{fgt}; \theta),
	\end{equation}
	where $x^{fgt}$ and $y^{fgt}$ respectively correspond to the prompts and responses in the forget data $D_{fgt}$. $L$ is the loss function, usually defined by Cross-Entropy loss.
	\begin{equation}
		\label{equ:CE_Loss}
		L_{CE} = -\frac{1}{K}\sum\nolimits_{i = 1}^{K}\sum\nolimits_{c = 1}^{C} y_{i,c} \cdot log(p_{i,c}),
	\end{equation}
	where $C$ denotes the size of the vocabulary, and $K$ denotes the total number of tokens in a sequence. $p_{i,c}$ is the probability of token $i$ belonging to class $c$. $y_{i,c}$ is $1$ if the actual token at position $i$ belongs to class $c$ and $0$ otherwise. GA forces the model to forget the target data by driving $p_{i,c}$ closer to 0 when $y_{i,c} = 1$.
	
	However, such inverse updates could irrevocably damage the model utility on the retain data due to catastrophic forgetting, in which a model forgets previously acquired knowledge or function during the process of learning new tasks. This occurs because unlearning requires removing specific data that may be integral to the model's effectiveness. To address this issue, one intuitive method named orthogonal gradient descent (OGD), derived from the field of continual learning, can be applied. OGD projects the gradients of new tasks $g$ onto a subspace orthogonal to that of previous tasks. Unfortunately, this approach still suffers from model utility reduction since the direction is not gradient descent for earlier tasks.
	
	Another way to counter this threat is to use KL divergence to constrain the unlearned model from diverging significantly from the original model. For instance, \cite{yao2023large, chen2023unlearn} incorporates a KL divergence term and weighted-sum with the $\mathcal{L}_{fgt}$, which aligns the output of the unlearned model with that of the original model on the retain data.
	
	However, a gradient conflict still arises from the simultaneous optimization of two partially opposing objectives: unlearning of certain knowledge and preservation of other essential information. That is, the gradient $g_{fgt}$ of the unlearning task and the gradient $g_{rt}$ of retain data preservation may exhibit an inner product $g_{fgt} \cdot g_{rt} < 0$. Consequently, the model update direction would easily conflict with $g_{fgt}$ and $g_{rt}$, leading to compromised effectiveness in both the process of unlearning and knowledge retention. We verify it in the experimental results of Table~\ref{tab:exps}.
	
	\section{Methodology}
	\subsection{Problem Formulation}
	To balance the trade-off between unlearning and model utility preservation, we formulate LLM unlearning as the following multi-objective optimization problem (MOP):
	\begin{equation}
		\label{Problem:MOO}
		\min\limits_{\theta} \{\mathcal{L}_{fgt}(\theta), \mathcal{L}_{KL}(\theta), \mathcal{L}_{rt}(\theta)\},
	\end{equation}
	where $\mathcal{L}_{KL}$ is the KL divergence between the outputs of the model and the original model on the retain data $D_{rt}$, which is formulated by \cite{yao2023large}:
	\begin{equation}
		\small
		\mathcal{L}_{KL}:= \sum_{(x^{rt}, y^{rt}) \in D_{rt}} \sum_{i=1}^{|y^{rt|}} \text{KL}\left(h_{\theta^0}(x^{rt}, y^{rt}_{<i}) \,\middle\|\, h_{\theta^t}(x^{rt}, y^{rt}_{<i})\right),
	\end{equation}
	where $\theta^0$ and $\theta^t$ correspond to the original model and the unlearned model at round $t$, $x^{rt}$ and $y^{rt}$ are non-harmful prompts and answers in the retain set, and $h_\theta(x, y_{<i}) := \mathbb{P}(y_i|(x, y_{<i}); \theta)$. It is used to maintain the model utility \cite{yao2023large}.
	
	$\mathcal{L}_{rt}$ represents the loss of maintaining the task performance, following the definition in \cite{chen2023unlearn}:
	\begin{equation}
		\mathcal{L}_{rt} := \sum_{(x^{rt}, y^{rt}) \in D_{rt}} L_{CE}(x^{rt}, y^{rt}; \theta).
	\end{equation}
	
\begin{figure}[t!]
		\centering
		\includegraphics[width=0.86\columnwidth]{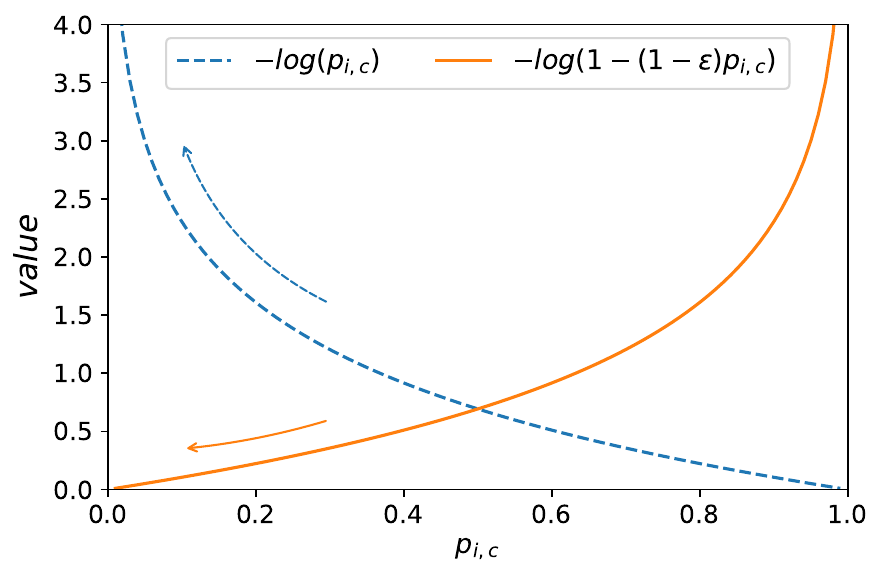}
		\caption{When using CE loss and GA for unlearning, $p_{i,c}$ is driven to 0, which leads to gradient explosion (blue line). As for UCE, it adopts the gradient descent to drive $p_{i,c}$ to 0 to unlearn, preventing gradient explosion.}
		\label{fig:uce_loss}
	\end{figure}

	$\mathcal{L}_{fgt}$ denotes the loss of unlearning. In GA, it takes the inverse of the CE loss (Eq.~(\ref{bad_loss})). However, since the Cross-Entropy loss has no upper bound, directly minimizing Eq.~(\ref{bad_loss}) would lead to gradient explosion, as visualized in Fig.~\ref{fig:uce_loss}.
	
	To address this issue, we design the following Unlearning Cross-Entropy (UCE) loss to replace the inverse CE loss (GA):
	\begin{equation}
		\label{equ:UCE_Loss}
		L_{UCE} = -\frac{1}{K}\sum\nolimits_{i = 1}^{K}\sum\nolimits_{c = 1}^{C} y_{i,c} \cdot log(1 - (1 - \epsilon) p_{i,c}),
	\end{equation}
	where $\epsilon$ is a small scalar used to slightly scale $p_{i,c}$ to prevent unbounded growth if $p_{i,c} = 1$ in the beginning of unlearning. Following this, $\mathcal{L}_{fgt}$ is defined by Eq.~\ref{eq:uce_loss}, enabling model updates via the common gradient descent method. Since the UCE loss has the lower bound 0, it can achieve the goal of unlearning without causing gradient explosion.
	\begin{equation}
		\label{eq:uce_loss}
		\mathcal{L}_{fgt} := \sum_{(x^{fgt}, y^{fgt}) \in D_{fgt}} L_{UCE}(x^{fgt}, y^{fgt}; \theta).
	\end{equation}
	\subsection{Dual Space Multiple Gradient Descent Algorithm}
	In the proposed MOLLM, we solve Problem (\ref{Problem:MOO}) by iterating $\theta^{t+1} = \theta^t + \eta^t d^t$ in each round $t$, where $\eta^t$ represents the step size (i.e., the learning rate), and $d^t$ denotes the model update direction, which is common descent for each sub-objective of Problem (\ref{Problem:MOO}). Specifically, Denote $g_{fgt}^t$, $g_{KL}^t$ and $g_{rt}^t$ as the gradients of the three sub-objective respective to the model parameters. $d^t$ satisfies $d^t$$\cdot$$g_{fgt}^t$$<$$0$, $d^t$$\cdot$$g_{KL}^t$$<$$0$, and $d^t$$\cdot g_{rt}^t$$<$$0$, allowing it to simultaneously minimize each sub-objective, thereby unlearning the forget data while preserving the model utility. Since $d^t$ is the common descent direction, based on the theorem of multiple gradient descent \cite{fliege2000steepest,FedMDFG,FedLF}, the model can reach the Pareto stationarity of the multi-objective optimization problem.
	
	\textbf{Definition 1 (Pareto Stationarity)}. $\theta^*$ is called Pareto stationary iff there exists a convex combination of $-g_{fgt}$, $-g_{KL}$, and $-g_{rt}$ that results in zero, i.e., $\exists \xi_i \geq 0, \sum_{i=1}^{3}=1$, such that $\xi_1 g_{fgt} + \xi_2 g_{KL} + \xi_3 g_{rt} = 0$.
	
	We propose a novel and efficient way to determine such a common descent direction $d^t$, named Dual Space Multiple Gradient Descent Algorithm (DS-MGDA). Let $S$ be the space spanned by $g_{fgt}$, $g_{KL}$, and $g_{rt}$. The key idea is to obtain the dual vectors of $g_{fgt}$, $g_{KL}$, $g_{rt}$, denoted as $g_{fgt}^*$, $g_{KL}^*$, $g_{rt}^*$, where the convex combinations of $g_{fgt}^*$, $g_{KL}^*$, $g_{rt}^*$ form the dual space of $S$, as seen in Fig.~\ref{fig:fig1}. Afterward, the common descent direction $d^t$ for the model update can be obtained by taking the average of $-g_{fgt}^*$, $-g_{KL}^*$, and $-g_{rt}^*$.
	
	\textbf{Definition 2 (Dual Space) \cite{paffenholz2010polyhedral}}. The space $S^*$ is called the dual space of $S$ iff for any vector $a$ lying in $S$, and for any $b$ lying in $S^*$, it has $a \cdot b < 0$.
	
	All vectors $d$ lying inside $S^*$ satisfy $d$$\cdot$$g_{fgt}^t$$<$$0$, $d$$\cdot$$g_{KL}^t$$<$$0$, and $d$$\cdot g_{rt}^t$$<$$0$. Since the dual vectors $g_{fgt}^*$, $g_{KL}^*$, $g_{rt}^*$ lie on the edge of $S^*$, one of them is orthogonal to the other two. Therefore, according to the matrix computation \cite{golub2013matrix}, $g_{fgt}^*$ can be obtained by projecting it to the null space of $g_{KL}^t$ and $g_{rt}^t$, following Eq.~(\ref{equ:proj_null_space}), and similarly for $g_{KL}^*$ and $g_{rt}^*$.
	\begin{equation}
		\label{equ:proj_null_space}
		g_{fgt}^* = g_{fgt}^t - A^T(AA^T)^{-1}Ag_{fgt}^t,
	\end{equation}
	where $A$ denotes a matrix concatenated by $g_{KL}^t$ and $g_{rt}^t$. 
	After obtaining $g_{fgt}^*$, $g_{KL}^*$, and $g_{rt}^*$, $d^t$ is obtained by $d^t = -\frac{1}{3}(g_{fgt}^* + g_{KL}^* + g_{rt}^*)$. We scale the length of $d^t$ to $\min \{\|g_{fgt}\|, \|g_{KL}\|, \|g_{rt}\|\}$ in the end because the length of the gradients can be increased by the projection.
	
	DS-MGDA is efficient, as it relies solely on fundamental matrix computations. Moreover, it is applicable to MOP with more than 3 objectives, provided that the matrix $A$ is full-rank.
	
	\begin{table*}[t]
		\centering
		\caption{Performance of Llama3-8B on PKU-SafeRLHF dataset after unlearning. The proposed MOLLM can eliminate the gradient conflicts during unlearning and achieve a lower harmful rate while preserving the model utility.}
		\label{tab:exps}
		\vspace{-7pt}
		\resizebox{0.89\linewidth}{!}{
			\begin{tabular}{llllllll}
				\toprule
				Method            & HR(\%) $\downarrow$ & Toxicity $\downarrow$ & Obscenity $\downarrow$ & Fluency $\downarrow$ & $PC_{fgt}$(\%) $\downarrow$ & $PC_{KL}$(\%) $\downarrow$ & $PC_{rt}$(\%) $\downarrow$ \\
				\midrule
				Original          & 79.18 & 0.095    & 0.034   & 1.754   & -         & -        & -        \\
				Re-fintuning      & 53.92 & 0.058    & 0.002   & 1.889   & -         & -        & -        \\
				Relabeling        & 5.72  & 0.028    & 0.005   & 1.943   & 44.27     & 58.60     & 57.58    \\
				GA + Weighted-sum & 5.46  & 0.019    & 0.004   & 1.883   & 19.87     & 22.32    & 0.17     \\
				GA + OGD          & 3.68  & 0.016    & 0.002   & 4.490    & 0         & 0        & 63.46    \\
				\midrule
				MOLLM (ours)             & 3.45  & \textbf{0.007}    & \textbf{0.002}   & \textbf{1.689}   & \textbf{0}         & \textbf{0}        & \textbf{0}        \\
				Ablation 1        & 3.57  & 0.014    & 0.003   & 4.266   & 0         & 0        & 0        \\
				Ablation 2        & 5.14  & 0.018    & 0.004   & 1.924   & 0         & 0        & 0        \\
				Ablation 3        & \textbf{3.38}  & 0.007    & 0.002   & 5.080    & 0         & 97.58    & 97.10      \\
				\bottomrule
			\end{tabular}
		}
		\vspace{-15pt}
	\end{table*}
	
	\section{Experiments}
	\subsection{Experimental Setup}
	We follow \cite{yao2023large} to conduct the experiments on the well-known dataset PKU-SafeRLHF \cite{ji2024beavertails}. It contains harmful questions and harmful \& non-harmful responses. We simulate a scenario in which an LLM is fine-tuned on a downstream dataset $D$, after which some harmful samples are identified from $D$ and need to be unlearned. The learning rate is selected from $\{1e-5, 5e-6, 1e-6\}$ with decay of 0.999 per round. We take the best performance of each method in comparison.
	
	We compare the proposed MOLLM with the following LLM unlearning methods: (1) GA-weighted-sum \cite{yao2023large}: based on GA, design the loss function as the weighted-sum of the unlearning loss, KL divergence, and the CE loss on the retain data. (2) GA-OGD \cite{OGD}: based on GA, project the unlearning gradient $g_{fgt}$ to a vector that is orthogonal to the subspace of $g_{KL}$ and $g_{rt}$. Besides, we also consider two other methods: re-finetuning the model from scratch on the retain data and the re-labeling approach \cite{relabeling}, which randomly rewrites the label of the forget set and finetunes the model on this modified dataset. We test the algorithm performance on the SOTA model LLama3-8b \cite{llama3}. Furthermore, we consider three variants of MOLLM to study the effect of each part: (1) Ablation 1: Change the UCE loss back to the CE loss, and take the reverse of the gradient similar to GA. (2) Ablation 2: The same as ablation 1, but with a smaller learning rate. (3) Ablation 3: Directly utilize $-g_{fgt}^t$ instead of the common descent direction to update the model. It's worth noting that, for GA-OGD and Ablation 1, we have to finetune the hyper-parameter of the gradient clipping to prevent the gradient explosion.
	
	\subsection{Evaluation Metrics}
	To evaluate the effectiveness of unlearning, we follow \cite{yao2023large} to measure the harmful rate (HR). We also use \textit{Toxicity} \cite{Detoxify} for toxicity evaluation, where the output of the LLM is fed into a specific evaluation model which determines the toxicity of the input. Besides, we measure the obscenity by \cite{Detoxify} in a similar way. To measure the model utility, we follow \cite{yao2023large} to compute \textit{fluency} \cite{su2022contrastive} of the model on the retain set, computed by $2^{test\text{ }loss}$. Furthermore, we calculate the probability of conflict between the model update direction and the $g_{fgt}^t$, $g_{KL}^t$, and $g_{rt}^t$ during the unlearning process, denoting these probabilities as $PC_{fgt}$, $PC_{KL}$, and $PC_{rt}$. 
	
	\subsection{Unlearning effectiveness and Model Utility}
	We list the comparison results of different algorithms in Table~\ref{tab:exps}. The row labeled `Original" denotes the metrics before unlearning. It can be seen that using GA and the weighted-sum methods result in poor unlearning performance, so does the relabeling method. Besides, finetuning the model on the retain set only reduces the harmful rate to 53.92\%. These results confirm the need for proactive unlearning. While GA-OGD demonstrates good unlearning effectiveness, it compromises model utility. This is because movement in the parameter space affects the orthogonality of the update direction to $g_{KL}^t$ and $g_{rt}$, so the orthogonal update could still increase $\mathcal{L}_{KL}$ and $\mathcal{L}_{rt}$. In comparison, MOLLM can achieve the lowest harmful rate, toxicity, and obscene value, while preserving the model utility. Moreover, the fluency value of MOLLM is decreased, benefiting from the multi-objective optimization of MOLLM.
	
	\subsection{Ablation Study}
	We compare different variants of MOLLM to inspect the effect of each component. As shown in Table~\ref{tab:exps}, switching from UCE loss to CE loss in Ablation 1 worsens unlearning performance and causes catastrophic forgetting, with fluency increasing to 4.266. This occurs because GA increases the gradient norm, affecting model convergence and potentially leading to gradient explosion. If we use a smaller learning rate (Ablation 2), the convergence is slower and has poorer results. This confirms that the unbounded CE loss makes hyper-parameter tuning difficult when using GA to unlearn, rendering it impractical. Ablation 3 suffers from catastrophic forgetting as fluency increases to 5.08, since the update direction $-g_{fgt}^t$  favors unlearning while ignoring model utility preservation.
	
	\section{Conclusion}
	In this paper, we study the proactive way of using Gradient Ascent for LLM unlearning, and analyze two significant challenges: gradient explosion and catastrophic forgetting. To address these issues, we incorporate multi-objective optimization into LLM unlearning, named MOLLM. Specifically, we design an unlearning version of Cross-Entropy loss to prevent gradient explosion, offering a novel and efficient way to calculate the common descent direction for the model update to unlearn the forget data while preserving the model utility. Experimental results demonstrate the effectiveness of MOLLM compared to the state-of-the-art baselines.

	\section*{Acknowledgment}
	This work is supported in part by National Natural Science Foundation of China (72331009, 72171206), in part by the Shenzhen Institute of Artificial Intelligence and Robotics for Society (AIRS), and in part by Shenzhen Key Lab of Crowd Intelligence Empowered Low-Carbon Energy Network (No. ZDSYS20220606100601002).
	
	\section*{Compliance with Ethical Standards}
	This research study was conducted using data made available in open access by (\url{https://github.com/PKU-Alignment/safe-rlhf}). Ethical approval was not required as confirmed by the license attached with the open access data.
	
	
	\bibliographystyle{IEEEtran} 
	\bibliography{refs.bib}

\end{document}